\documentclass[10pt,twocolumn,letterpaper]{article}
\pdfoutput=1
\usepackage{cvpr}
\usepackage{times}
\usepackage{epsfig}
\usepackage{graphicx}
\usepackage{amsmath}
\usepackage{amssymb}
\usepackage{multirow}
\usepackage{booktabs}
\usepackage{gensymb}

\usepackage[pagebackref=true,breaklinks=true,letterpaper=true,colorlinks,bookmarks=false]{hyperref}
\usepackage{xcolor}
\usepackage{bbm}

\cvprfinalcopy 
\newcommand{\todo}[1]{{\color{red}{TODO: [#1]}}}

\newcommand*{\Cdot}{\raisebox{-0.25ex}{\scalebox{1.75}{$\cdot$}}}
\newcommand{\mypara}{\vspace*{-3mm}\paragraph}

\newcommand{\myitem}{\vspace*{-2mm}\item[\Cdot]}
\newcommand{\namelong}{Articulation-aware  Normalized Coordinate Space Hierarchy\xspace}
\newcommand{\nameshort}{ANCSH\xspace}

\def\cvprPaperID{1891} 

\definecolor{mygreen}{rgb}{0.3,0.9,0.4}
\definecolor{newgreen}{rgb}{0.1,0.9,0.5}

\newcommand\blfootnote[1]{%
  \begingroup
  \renewcommand\thefootnote{}\footnote{#1}%
  \addtocounter{footnote}{-1}%
  \endgroup
}
\begin{document}


\title{Category-Level Articulated Object Pose Estimation \vspace{-3mm}}
\author{
Xiaolong Li$^{1\ast}$\quad
He Wang$^{2\ast}$\quad
Li Yi$^{3}$ \quad
Leonidas Guibas $^{2}$ \quad
A. Lynn Abbott $^{1}$ \quad
Shuran Song$^{4}$
\vspace{1mm}\\ 
$^{1}$Virginia Tech \quad
$^{2}$Stanford University \quad
$^{3}$Google Research\quad
$^{4}$Columbia University\quad \vspace{1mm} \\
\href{https://articulated-pose.github.io/}{articulated-pose.github.io}
\vspace{-3mm} 
}

\maketitle

\begin{abstract}
This paper addresses the task of category-level pose estimation for articulated objects from a single depth image.
We present a novel category-level approach that correctly accommodates object instances previously unseen during training. We introduce \namelong (\nameshort) -- a canonical representation for different articulated objects in a given category.
As the key to achieve intra-category generalization, the representation constructs a canonical object space as well as a set of canonical part spaces. The canonical object space normalizes the object orientation, scales and articulations (e.g. joint parameters and states) while each canonical part space further normalizes its part pose and scale.
We develop a deep network based on PointNet++ that predicts \nameshort from a single depth point cloud, including part segmentation, normalized coordinates, and joint parameters in the canonical object space. 
By leveraging the canonicalized joints, we demonstrate: 
1) improved performance in part pose and scale estimations using the induced kinematic constraints from joints; 2) high accuracy for joint parameter estimation in camera space.
\end{abstract}
\vspace{-5mm} 
        


\section{Introduction}
\begin{figure}[t]
    \centering
    \includegraphics[width=\linewidth]{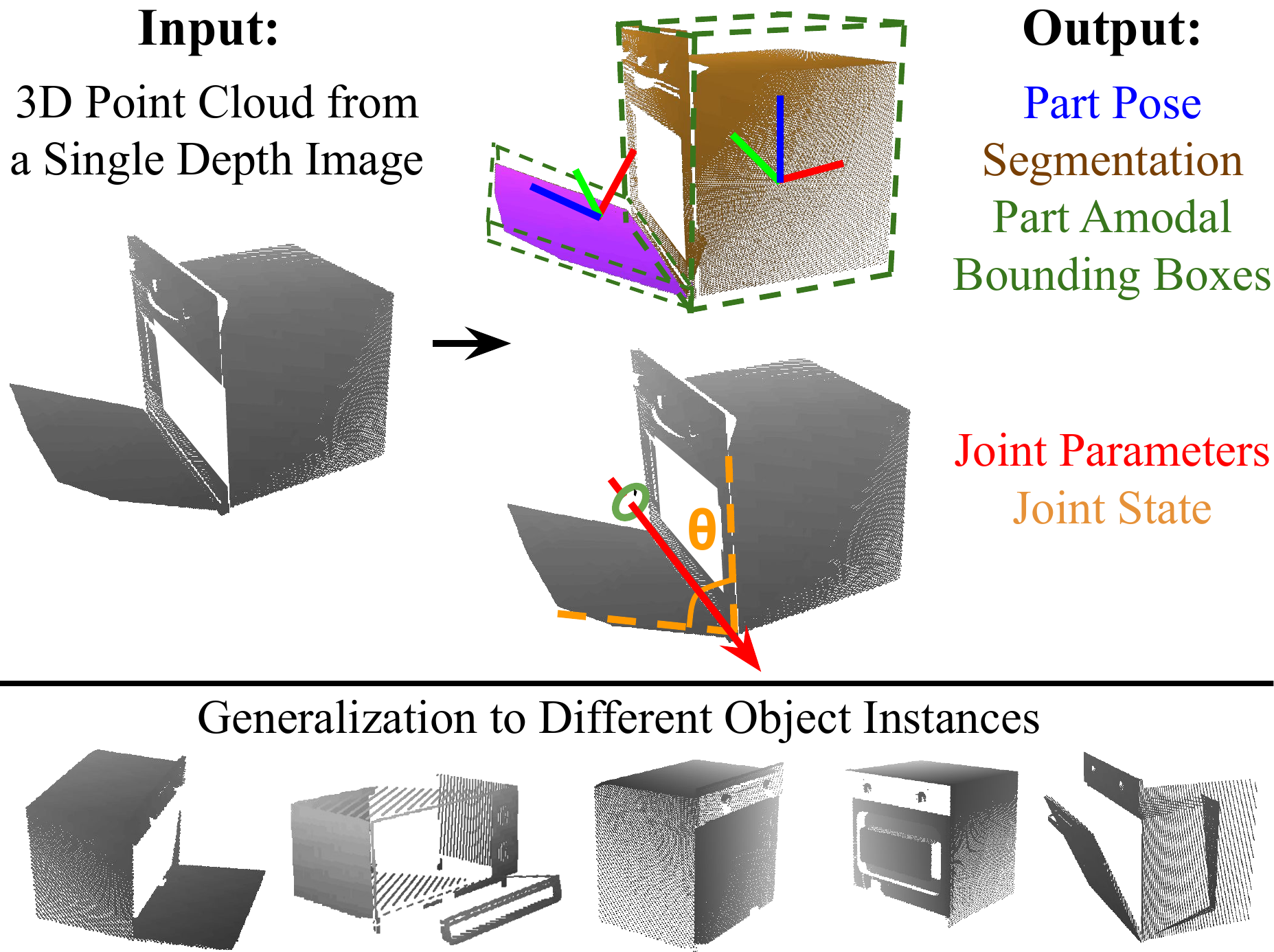}
    \caption{\textbf{Category-level articulated object pose estimation.} 
    Given a depth point cloud of a novel articulated object from a known category, our algorithm estimates: part attributes, including part segmentation, poses, scales and amodal bounding boxes; joint attributes, including joint parameters and joint states.
    }
    \label{fig:task}
    \vspace{-5mm}
\end{figure}
\blfootnote{* indicates equal contributions.}
Our environment is populated with articulated objects, ranging from furniture such as cabinets and ovens to small tabletop objects such as laptops and eyeglasses. 
Effectively interacting with these objects requires a detailed understanding of their articulation states and part-level poses. Such  understanding is beyond the scope of typical 6D pose estimation algorithms, which have been designed for rigid objects \cite{xiang2017posecnn, tremblay2018deep, sundermeyer2018implicit, wang2019normalized}. 
Algorithms that do consider object articulations \cite{katz2008manipulating, katz2013interactive, hausman2015active, martin2016integrated} often require the exact object CAD model and the associated joint parameters at test time, preventing them from generalizing to new object instances. 

In this paper, we adopt a learning-based approach to perform category-level pose estimation for articulated objects. Specifically, we consider the task of estimating per-part 6D poses and 3D scales, joint parameters (i.e. type, position, axis orientation), and joint states (i.e. joint angle) of a novel articulated object instance in a known category from a single depth image. Here object instances from one category will share a known kinematic chain composing of a fixed number of rigid parts connected by certain types of joints. We are particularly interested in the two most common joint types, revolute joints that cause 1D rotational motion (\eg, door hinges), and prismatic joints that allow 1D translational movement (\eg, drawers in a cabinet). An overview is shown in  Figure \ref{fig:task}.
To achieve this goal, several major challenges need to be addressed:

First, to handle novel articulated objects without knowing exact 3D CAD models, we need to find a shared representation for different instances within a given category. 
The representation needs to accommodate the large variations in part geometry, joint parameters, joint states, and self-occlusion patterns. More importantly, for learning on such diverse data, the representation needs to facilitate intra-category generalization.

Second, in contrast to a rigid object, an articulated object is composed of multiple rigid parts leading to a higher degree of freedom in its pose. 
Moreover, the parts are connected and constrained by certain joints and hence their poses are not independent.
It is challenging to accurately estimate poses in such a high-dimensional space while complying with physical constraints.

Third, various types of joints
provide different physical constraints and priors for part articulations.
Designing a framework that can accurately predict the parameters and effectively leverage the constraints for both revolute and prismatic joints is yet an open research problem. 

To address the representation challenge, we propose a shared category-level representation for different articulated object instances, namely, \namelong (\nameshort). 
Concretely, \nameshort is a two-level hierarchy of canonical space, composed of Normalized Articulated Object Coordinate Space (NAOCS) at the root level and a set of Normalized Part Coordiante Spaces (NPCSs) at the leaf level.
In the NAOCS, object scales, orientations, and joint states are normalized. In the NPCS of each rigid part, the part pose and scale are further normalized . We note that NAOCS and NPCS are complimentary to each other: NAOCS provides a canonical reference on the object level while NPCSs provide canonical part references. The two-level reference frames from \nameshort allow us to define per-part pose as well as joint attributes for previously unseen articulated object instances on the category-level.

To address the pose estimation challenge, we segment objects into multiple rigid parts and predict the normalized coordinates in \nameshort. 
However, separate per-part pose estimation could easily lead to physically impossible solutions since joint constraints are not considered. To conform with the kinematic constraints introduced by joints, 
we estimate joint parameters in the NAOCS from the observation, mathematically model the constraints based upon the joint type, and then leverage the kinematic priors to regularize the part poses.
We formulate articulated pose fitting from the \nameshort to the depth observation as a combined optimization problem, taking both part pose fitting and joint constraints into consideration. In this work we mainly focus on 1D revolute joints and 1D prismatic joints, while the above formulation can be extended to model and support other types of joints.

Our experiments demonstrate that leveraging the joint constraints in the combined optimization leads to improved performance in part pose and scale prediction. 
Noting that leveraging joint constraints for regularizing part poses requires high-accuracy joint parameter predictions, which itself is very challenging. Instead of directly predicting joint parameters in the camera space, we consider and leverage predictions in NAOCS, where joints are posed in a canonical orientation, e.g. the revolute joints always point upward for eyeglasses. By transforming joint parameter predictions from NAOCS back to camera space, we further demonstrate supreme accuracy on camera-space joint parameter predictions.
In summary, the primary contribution of our paper is a unified framework for category-level articulated pose estimation. 
In support of this framework, we design: 
\begin{itemize}
    \myitem A novel  category-level representation for articulated objects -- \namelong (\nameshort).
    \myitem A PointNet++ based neural network that is capable of predicting \nameshort for previously unseen articulated object instances from a single depth input. 
    \myitem A combined optimization scheme that leverages \nameshort predictions along with induced joint constraints for part pose and scale estimation.
    \myitem A two-step approach for high-accuracy joint parameter estimation that first predicts joint parameters in the NAOCS and then transforms them into camera space using part poses.
\end{itemize}

\section{Related Work}
This section summarizes related work on pose estimation for rigid  and articulated objects.

\mypara{Rigid object pose estimation.}
Classically, the goal of  pose estimation is to infer an object's 6D pose (3D rotation and 3D location) relative to a given reference frame. Most  previous work has  focused on estimating instance-level pose 
by assuming that exact 3D CAD models are available. For example, traditional 
algorithms such as iterative closest point (ICP)~\cite{besl1992method} perform template matching by aligning the CAD model with an observed 3D point cloud. Another family of approaches 
aim to regress the object coordinates onto its CAD model for each observed object pixel, and then use voting to solve for object pose~\cite{brachmann2014learning, brachmann2016uncertainty}. These approaches are limited by the need to have exact CAD models for particular object instances. 

Category-level pose estimation aims to infer an object's pose and scale relative to a category-specific canonical representation. Recently, Wang \etal \cite{wang2019normalized} extended the object coordinate based approach to perform category-level pose estimation. The key idea behind the intra-category generalization is to regress the coordinates within a Normalized Object Coordinate Space (NOCS), where the sizes are normalized and the orientations are aligned for objects in a given category. Whereas the work by \cite{wang2019normalized} 
focuses on pose and size estimation for rigid objects, the work presented here extends the NOCS concept to accommodate articulated objects at both part and object level. 
In addition to pose, our work also infers joint information and addresses particular problems related to occlusion.

\begin{figure*}[t]
    \centering
     \vspace{-3mm}
    \includegraphics[width=\linewidth]{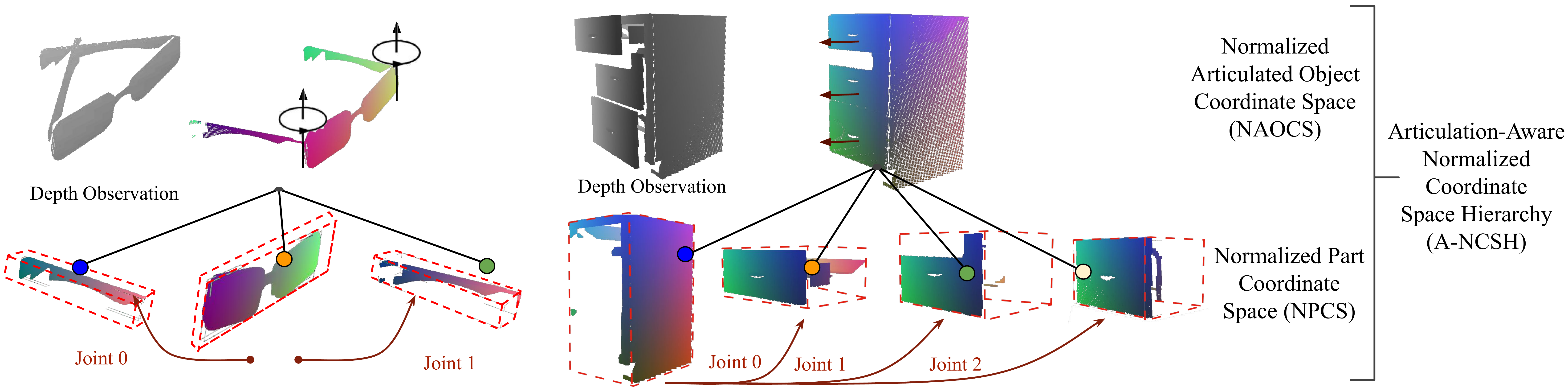}
    \caption{ \textbf{ Articulation-aware  Normalized Coordinate  Space  Hierarchy (ANCSH)} is a category-level  object representation composed of a Normalized Articulated Object Coordinate Space (NAOCS) on top of a set of Normalized Part  Coordinate  Spaces (NPCSs) per part. Here we show two examples of \nameshort representation (points are colored according to its corresponding coordinates in the NAOCS/NPCS). Note that NAOCS sets the object articulations to pre-defined states, all the joints in the NAOCS are hence canonicalized, e.g. the axes of the revolute joints in the eyeglasses example all point upwards and the joint angles are right angles. For each individual part, NPCS maintains the part orientation as in the NAOCS but zero-centers its position and normalizes its scales.}
    \label{fig:rep}
    \vspace{-3mm}
\end{figure*}

\mypara{Articulated object pose estimation. }
Most algorithms that attempt pose estimation for articulated objects assume that  instance-level information is available. The approaches often use  CAD models for particular instances along with known kinematic parameters to constrain  the search space and to recover the pose separately for different parts \cite{michel2015pose, desingh2018factored}. Michel \etal \cite{michel2015pose} used a random forest  to vote for pose parameters on canonical body parts for each point in a depth image, followed by a variant of the Kabsch algorithm to estimate  joint parameters using RANSAC-based energy minimization. 
Desingh \etal \cite{desingh2018factored} adopted a generative approach using a Markov Random Field formulation,  factoring the state as individual parts constrained by their articulation parameters. However, these approaches only consider known object instances and cannot handle different part and kinematic variations. A recent work \cite{abbatematteo2019learning} also tries to handle novel objects within the same category by training a mixed density model \cite{bishop1994mixture} on depth images, their method could infer kinematic model using probabilities predictions of a mixtures of Gaussians. However they don't explicitly estimate pose on part-level, the simplified geometry predictions like length, width are for the whole object with scale variation only. 

Another line of work relies on active manipulation of an object to infer its articulation pattern \cite{katz2008manipulating, katz2013interactive, hausman2015active, martin2016integrated, yi2018deep}. For example, Katz \etal \cite{katz2013interactive}, uses a robot manipulator to interact with articulated objects as RGB-D videos are recorded. Then the 3D points are clustered into rigid parts according to their motion.
Although these approaches could perform pose estimation for unknown objects, they require the input to be a sequence of images that observe an object's different articulation states, whereas our approach is able to perform the task using a single depth observation.

\mypara{Human body and hand pose estimation.}
Two specific articulated classes have gained considerable attention recently: the human body and the human hand. 
For human pose estimation, approaches have been developed using end-to-end networks to  predict 3D joint locations directly~\cite{mehta2017vnect, sun2017compositional,pavlakos2017coarse}, using dense correspondence maps between 2D images and 3D surface models~\cite{alp2018densepose}, or estimating full 3D shape  through 2D supervision~\cite{lassner2017unite, pavlakos2018learning}. 
%
Similarly, techniques for hand pose estimation (\eg, \cite{wan2018dense, ge2018point})  leverages dense coordinate regression, which is then used for voting 3D joint locations. 
Approaches for both body and hand pose estimation are often specifically customized for those object types, relying on a fixed skeletal model with class-dependent variability (\eg, expected joint lengths) and strong shape priors (\eg, using parametric body shape model for low-dimensional parameterization).
Also, such hand/body approaches accommodate only revolute joints. In contrast, our algorithm is designed to handle generic articulated objects with varying kinematic chain,  allowing both revolute joints and prismatic joints. 


\section{Problem Statement}
\vspace{-3mm}
The input to the system is a 3D point cloud $P={\{\mathbf{p}_i\in \mathbb{R}^3|~i=1,...,N\}}$ backprojected from a single depth image
representing an unknown object instance from a known category, where $N$ denotes the number of points. 
We know that all objects from this category share the same kinematic chain composed of $M$ rigid parts $\{S^{(j)}~|~j=1,...,M\}$ and $K$ joints with known types $\{J_k~|~k=1,...,K\}$. 
The goal 
is to segment the point cloud into rigid parts $\{S^{(j)}\}$, recover the 3D rotations $\{R^{(j)}\}$, 3D translations $\{\mathbf{t}^{(j)}\}$, and sizes $\{s^{(j)}\}$ for the parts in $\{S^{(j)}\}$, and predict the joint parameter $\{\phi_k\}$ and state $\{\theta_k\}$ for the joints in $\{J_k\}$.
In this work, we consider 1D revolute joints and 1D prismatic joints. We parameterize the two types of joints as following. For a revolute joint, its joint parameters include the direction of the rotation axis $\mathbf{u}_k^{(r)}$ as well as a pivot point $\mathbf{q}_k$ on the rotation axis; its joint state is defined as the relative rotation angle along $\mathbf{u}_k^{(r)}$ between the two connected parts compared with a pre-defined rest state. For a prismatic joint, its joint parameter is the direction of the translation axis $\mathbf{u}_k^{(t)}$, and its joint state is defined as the relative translation distance along $\mathbf{u}_k^{(t)}$ between the two connected parts compared with a pre-defined rest state. 

\section{Method}

\nameshort provides a category-specific reference frame defining per-part poses as well as joint attributes for previously unseen articulated object instances.
In Sec.~\ref{sec:representation}, we first explain \nameshort in detail. In Sec.~\ref{sec:network}, we then present a deep neural network capable of predicting the \nameshort representation. Sec.~\ref{sec:optimization} describes how the \nameshort representation is used to jointly optimize part poses with explicit joint constraints.
Last, we describe how we compute joint states and deduce camera-space joint parameters in Sec.~\ref{sec:joint}.

\subsection{\nameshort Representation}
\label{sec:representation}
Our \nameshort representation is inspired by and closely related to Normalized Object Coordinate Space (NOCS) \cite{wang2019normalized}, which we briefly review here. NOCS is defined as a 3D space contained within a unit cube and was introduced in \cite{wang2019normalized} to estimate the category-level 6D pose and size of rigid objects. For a given category, the objects are consistently aligned by their orientations in the NOCS. Furthermore, these objects are zero-centered and uniformly scaled so that their tight bounding boxes are all centered at the origin of the NOCS with a diagonal length of 1. NOCS provides a reference frame for rigid objects in a given category so that the object pose and size can then be defined using the similarity transformation from the NOCS to the camera space. 
However, NOCS is limited for representing articulated objects. Instead of the object pose and size, we care more about the poses and the states for each individual parts and joints, which isn't addressed in NOCS. 

To define category-level per-part poses and joint attributes, we present \nameshort, a two-level hierarchy of normalized coordinate spaces, as shown in Figure \ref{fig:rep}.
At  the  root level, NAOCS provides an object-level reference frame with normalized pose, scale, and articulation; at the leaf level, NPCS provides a reference frame for each individual part.
We explain both NPCS and NAOCS in detail below.

\mypara{NAOCS.}
To construct a category-level object reference frame for the collection of objects, we first bring all the object articulations into a set of pre-defined rest states. Basically, for each joint $J_k$, we manually define its rest state $\theta_{k0}$ and then set the joint into this state. For example, we define the rest states of the two revolute joints in the eyeglasses category to be in right angles; we define the rest states of all drawers to be closed. In addition to normalizing the articulations, NAOCS applies the same normalization used in \cite{wang2019normalized} to the objects, including zero-centering, aligning orientations, and uniformly scaling.  

As a canonical object representation, NAOCS has the following advantages: 
1) the joints are set to predefined states so that accurately estimating joint parameters in NAOCS, e.g. the direction of rotation/translation axis, becomes an easy task;
2) with the canonical joints, we can build simple mathematical models to describe the kinematic constraints regarding each individual joint in NAOCS.

\mypara{NPCS.}
For each part, NPCS further zero-centers its position and uniformly scales it as is done in \cite{wang2019normalized}, while at the same time keeps its orientation unchanged as in NAOCS. In this respect, NPCS is defined similarly to NOCS \cite{wang2019normalized} but for individual parts instead of whole objects. NPCS provides a part reference frame and we can define the part pose and scale as the transformation from NPCS to the camera space. Note that corresponding parts of different object instances are aligned in NPCS, which facilitates intra-category generalization and enables predictions for unseen instances.


\mypara{Relationship between NPCS, NAOCS and NOCS.}
Both NPCS and NAOCS are inspired by the NOCS representation and designed for handling a collection of articulated objects from a given category. Therefore, similar to NOCS, both representations encode canonical information and enable generalization to new object instances. However, each of the two representations has its own advantages in modeling articulated objects and hence provides complementary information. Thus, our \nameshort leverages both NPCS and NAOCS to form a comprehensive representation of both parts and articulations. 

On the one hand, NPCSs normalize the position, orientation, and size for each part. Therefore, transformation between NPCSs and camera space can naturally be used to compute per-part 3D amodal bounding boxes, which is not well-presented in NAOCS representation. 
On the other hand, NAOCS looks at the parts from a holistic view, encoding the canonical relationship of different parts in the object space. NAOCS provides a parent reference frame to those in NPCSs and allows a consistent definition of the joint parameters across different parts. We hence model joints and predict joint parameters in the NAOCS instead of NPCSs. The joint parameters can be used to deduce joint constraints, which can regularize the poses between connected parts. 
Note that the information defined in NPCS and NAOCS is not mutually exclusive -- each NPCS can transform into its counterpart in NAOCS by a uniform scaling and translation. Therefore, instead of independently predicting the full NAOCS representation, our network predicts the scaling and translation parameters for each object part and directly applies it on the corresponding NPCS to obtain the NAOCS estimation.

\begin{figure}[t]
    \centering
    \includegraphics[width=\linewidth]{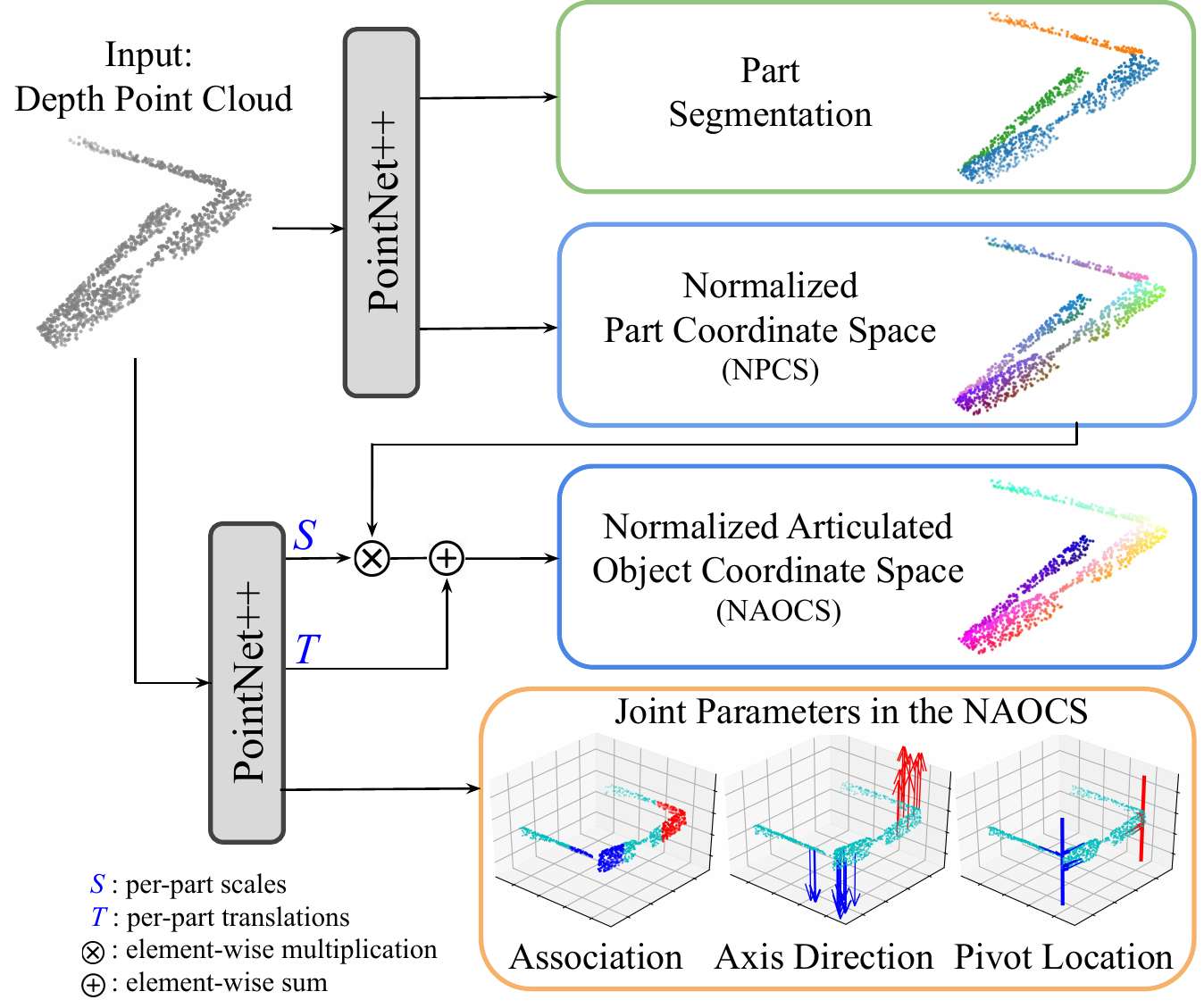}
    \caption{\textbf{\nameshort network} leverages two PointNet++~\cite{qi2017pointnet++} modules to predict the \nameshort representation, including part segmentation, NPCS coordinates, transformations (1D scaling and 3D translation) from  each  NPCS  to the NAOCS,  and  joint  parameters in the NAOCS.  This figure illustrates the eyeglasses case with only revolute joints, but the network structure also applies to objects with revolute and prismatic joints.
    }
    \label{fig:overview}
     \vspace{-5mm}
\end{figure}

\subsection{\nameshort Network}
\label{sec:network}
We devise a deep neural network capable of predicting the \nameshort representation for unseen articulated object instances. As  shown in Figure~\ref{fig:overview}, the network takes a depth point cloud $P$ as input and its four heads output rigid part segmentation, dense coordinate predictions in each NPCS, transformations from each NPCS to NAOCS, and joint parameters in NAOCS, correspondingly. The network is based on two modules adapted from the PointNet++~\cite{qi2017pointnet++} segmentation architecture.

The part segmentation head predicts a per-point probability distribution among the $M$ rigid parts. The NPCS head predicts $M$ coordinates ${\{\mathbf{c}_i^{(j)}\in\mathbb{R}^3|~j=1, ..., M\}}$ for each point $\mathbf{p}_i$. We use the predicted part label to select the corresponding NPCS. This design helps to inject the geometry prior of each part into the network and hence specializes the networks on part-specific predictions.
We design the segmentation network and the NPCS regression network to share the same PointNet++ backbone and only branch at the last fully-connected layers.

The NAOCS head predicts the transformations $\{G^{(j)}\}$ from each NPCS to the NAOCS, and computes the coordinates in NAOCS using the predicted transformations. 
Since part orientations are the same between NPCS and NAOCS, the network only needs to estimate a 3D translation $G_t^{(j)}$ and a 1D scaling $G_s^{(j)}$ for the NPCS of the part $S^{(j)}$. 
Similar to NPCS head, the head here predicts for each point $\mathbf{p}_i$ dense transformations with $G_{t, i}^{(j)}$ and $G_{s, i}^{(j)}$ for each NPCS of the parts $S^{(j)}$.
We use the predicted segmentation label to select per-point translation $G_{t,i}$ and scaling $G_{s,i}$.
Then the NAOCS coordinates can be represented as ${\{\textbf{g}_i|~\textbf{g}_i=G_{s,i}\textbf{c}_i+G_{t,i}\}}$. 
Finally, we compute $G_s^{(j)}$ and $G_t^{(j)}$ by averaging over points $\{\mathbf{p}_i \in S^{(j)}\}$.

The last head infers joint parameters $\{\phi_k^\prime\}$ for each joint $J_k$ in the NAOCS space (we use ``\,$^\prime$\,'' here to distinguish the NAOCS parameters from camera-space parameters.) We consider the following two types of joints: 1D revolute joint whose parameters include the rotation axis direction and the pivot point position, namely $\phi_{k}^\prime=(\mathbf{u}_{k}^{(r)\prime}, \mathbf{q}_{k}^\prime)$; 1D prismatic joint whose parameters is the translation axis direction $\phi_{k}^\prime=(\mathbf{u}_{k}^{(t)\prime})$.
We adopt a voting scheme to accurately predict joint parameters, in which we first associate points to each joint via a labeling scheme and then let the points vote for the parameters of its associated joint. 

We define a per-point joint association $\{a_i~|~ a_i\in \{0, 1, ..., K\}\}$, where label $k$ means the point $\mathbf{p}_i$ is associated to the joint $J_k$ and label $0$ means no association to any joint. We use the following heuristics to provide the ground truth joint association: for a revolute joint $J_k$, if a point $\mathbf{p}_i$ belongs to its two connecting parts and is within a distance $\sigma$ from its rotation axis, then we set $a_i = k$; for a prismatic joint, we associate it with all the points on its corresponding moving part. We empirically find $\sigma = 0.2$ leads to a non-overlapping joint association on our data. 

In addition to predicting joint association, the joint parameter head performs dense regression on the associated joint parameters. To be more specific, for each point $\mathbf{p}_i$, the head regresses a 7D vector $\mathbf{v}_i\in \mathbb{R}^7$. The first three dimensions of $\mathbf{v}_i$ is a unit vector, which either represents $\mathbf{u}^{(r)\prime}$ for a revolute joint or $\mathbf{u}^{(t)\prime}$ for a prismatic joint. The rest four dimensions are dedicated to the pivot point $\mathbf{q}^\prime$ in case the point is associated to a revolute joint. Since the pivot point of a 1D revolute joint is not uniquely defined (it can move arbitrarily along the rotation axis), we instead predict the projection of $\mathbf{p}_i$ to the rotation axis of its associated revolute joint by regressing a 3D unit vector for the projection direction and a scalar for the projection distance. For training, we only supervise the matched dimensions of $\mathbf{v}_i$ for points $\mathbf{p}_i$ with $a_i \neq 0$. We use the ground truth joint parameters $\phi_{a_i}^\prime$ associated with joint $J_{a_i}$ as the supervision. During inference, we use the predicted joint association to interpret $\mathbf{v}_i$. We perform a voting step to get the final joint parameter prediction $\phi_k^\prime$, where we simply average the predictions from points associated with each joint $J_k$. Note that the NAOCS head and the joint parameter head share the second PointNet++ as their backbone since they all predict attributes in the NAOCS.
\mypara{Loss functions:} We use relaxed IoU loss \cite{yi2018deep} $L_{\text{seg}}$ for part segmentation as well as for joint association $L_{\text{association}}$. We use mean-square loss $L_{\text{NPCS}}$ for NPCS coordinate regression. We use mean-square loss $L_{\text{NAOCS}}$ for NAOCS to supervise per-point translation $\{G_{t,i}^{(j)}\}_{i, j}$ and scaling $\{G_{s,i}^{(j)}\}_{i,j}$. We again use mean-square loss $L_{\text{joint}}$ for joint parameter predictions. Our total loss is given by $L = {\lambda_{1} L_{\text{seg}} + \lambda_{2} L_{\text{NPCS}} + \lambda_{3} L_{\text{NAOCS}} + \lambda_4 L_{\text{association}}} + \lambda_5 L_{\text{joint}}$, where the loss weights are set to $[1, 10, 1, 1, 1]$.



\subsection{Pose Optimization with Kinematic Constraints}
\label{sec:optimization}
Given the output of our \nameshort network,  including part segmentation, $\{\textbf{c}_i\}$ for each point $\mathbf{p}_i$, $\{ G_{t}^{(j)}, G_{s}^{(j)}\}$ for each part $S^{(j)}$, and $\{\phi_k^\prime\}$ for each joint $J_k$,  we now estimate the 6D poses and sizes $\{R^{(j)}, \textbf{t}^{(j)}, s^{(j)}\}$ for each part $S^{(j)}$.

Considering a part $S^{(j)}$, for the points $\{\mathbf{p}_i \in S^{(j)}\}$, we have their corresponding NPCS predictions $\{\mathbf{c}_i|\mathbf{p}_i \in S^{(j)}\}$. We could follow \cite{wang2019normalized} to perform pose fitting, where the Umeyama algorithm \cite{umeyama1991least} is adopted within a RANSAC \cite{fischler1981random} framework to robustly estimate the 6D pose and size of a single rigid object. However, without leveraging joint constraints, naively applying this approach to each individual part in our setting would easily lead to physically impossible part poses. 
To cope with this issue, we propose the following optimization scheme leveraging kinematic constraints for estimating the part poses. Without the kinematic constraints, the energy function $E_{\text{vanilla}}$ regarding all part poses can be written as $E_{\text{vanilla}}=\sum_j e_j$, where $$ e_j=\frac{1}{|S^{(j)}|}\sum_{\textbf{p}_i\in S^{(j)}}||\textbf{p}_i - (s^{(j)}R^{(j)}\textbf{c}_i+\textbf{t}^{(j)})||^2$$
We then introduce the kinematic constraints by adding an energy term $e_k$ for each joint to the energy function. In concrete terms, our modified energy function is $E_{\text{constrained}}=\sum_j e_j + \lambda\sum_k e_k$, where $e_k$ is defined differently for each type of joint. For a revolute joint $J_k$ with parameters $\phi_k^\prime=(\textbf{u}_k^{(r)\prime}, \textbf{q}_k^\prime)$ in the NAOCS, assuming it connects part $S^{(j_1)}$ and part $S^{(j_2)}$, we define $e_k$ as:
\begin{align*}
    e_k = ||R^{(j_1)}\textbf{u}_k^{(r)\prime}-R^{(j_2)}\textbf{u}_k^{(r)\prime}||^2
\end{align*}
For a prismatic joint $J_k$ with parameters $\phi_k^\prime=(\textbf{u}_k^{(t)\prime})$ in the NAOCS, again assuming it connects part $S^{(j_1)}$ and part $S^{(j_2)}$, we define $e_k$ as:
\begin{align*}
    e_k = \mu || R^{(j_1)}R^{(j_2)~T}-I ||^2 + \sum_{j=j_1,j_2}||[R^{(j)}\mathbf{u}_k^{(t)\prime}]_{\times}\delta_{j_1,j_2}||^2
\end{align*}
where $[\cdot]_{\times}$ converts a vector into a matrix for conducting cross product with other vectors, and $\delta_{j_1,j_2}$ is defined as:
\begin{align*}
    \delta_{j_1,j_2} = \textbf{t}^{(j_2)}-\textbf{t}^{(j_1)}+s^{(j_1)}R^{(j_1)}G_t^{(j_1)}-s^{(j_2)}R^{(j_2)}G_t^{(j_2)}
\end{align*}

To minimize our energy function $E_{\text{constrained}}$, we can no longer separately solve different part poses using the Umeyama algorithm. Instead, we first minimize $E_{\text{vanilla}}$ using the Umeyama algorithm to initialize our estimation of the part poses. Then we fix $\{s^{(j)}\}$ and adopt a non-linear least-squares solver to further optimize $\{R^{(j)}, \textbf{t}^{(j)}\}$, as is commonly done for bundle adjustment \cite{agarwal2010bundle}. Similar to \cite{wang2019normalized}, we also use RANSAC for outlier removal. 

Finally, for each part $S^{(j)}$, we use the fitted $R^{(j)}, \textbf{t}^{(j)}, s^{(j)}$ and the NPCS $\{\mathbf{c_i} | \mathbf{p}_i \in S^{(j)}\}$ to compute an amodal bounding box, the same as in \cite{wang2019normalized}.


\subsection{Camera-Space Joint Parameters and Joint States Estimation}
\label{sec:joint}
Knowing $\{R^{(j)}, \textbf{t}^{(j)}, s^{(j)}, G_{t}^{(j)}, G_{s}^{(j)}\}$ of each part, we can compute the joint states $\{\theta_k\}$ and deduce joint parameters $\{\phi_k\}$ in the camera space from NAOCS joint parameters $\{\phi_k^\prime\}$. For a revolute joint $J_k$ connecting parts $S^{(j_1)}$ and $S^{(j_2)}$, we compute its parameters $\phi_k=(\textbf{u}_k^{(r)}, \textbf{q}_k)$ in the camera space as:
\begin{align*}
    & \textbf{u}_k^{(r)} = \frac{(R^{(j_1)}+R^{(j_2)})\textbf{u}_k^{(r)\prime}}{||(R^{(j_1)}+R^{(j_2)})\textbf{u}_k^{(r)\prime}||}\\
    & \textbf{q}_k = \frac{1}{2}\sum_{j=j_1,j_2} \frac{R^{(j)}s^{(j)}}{G_s^{(j)}}\left(\textbf{q}_k^\prime-G_t^{(j)}\right)+\textbf{t}^{(j)}
\end{align*}
The joint state $\theta_k$ can be computed as:
\begin{align*}
    \theta_k=\text{arccos}((\text{trace}(R^{(j_2)}(R^{(j_1)})^T)-1)/2)
\end{align*}
For a prismatic joint $J_k$ connecting parts $S^{(k1)}$ and $S^{(k2)}$, we compute its parameters $\phi_k=(\textbf{u}_k^{(t)})$ in the camera space similar to computing $\textbf{u}_k^{(r)}$ for revolute joints and and its state $\theta_k$ is simply $||\delta_{k1,k2}||$.

\section{Evaluation}
\begin{table*}[t]
{
\footnotesize
\centering
\setlength\tabcolsep{4.2px}
\begin{tabular}{c|c|c|c|c|c|c|c}
\toprule
\raisebox{\dimexpr-.5\normalbaselineskip-1\cmidrulewidth-.5\aboverulesep}[0pt][0pt]{Category}                     & \raisebox{\dimexpr-.5\normalbaselineskip-1\cmidrulewidth-.5\aboverulesep}[0pt][0pt]{Method} & \multicolumn{3}{c|}{Part-based Metrics}                    & \multicolumn{1}{c|}{Joint States} & \multicolumn{2}{c}{Joint Parameters}  \\\cmidrule{3-8}
                             &        & Rotation Error $\downarrow$    &  Translation  Error  $\downarrow$      & 3D IoU $\%$ $\uparrow$         &  Error  $\downarrow$                              & Angle error  $\downarrow$          & Distance error  $\downarrow$       \\ \midrule
\multirow{2}{*}{Eye- }        &   NPCS & 4.0$\degree $, 7.7$\degree $, 7.2$\degree $      & 0.044, 0.080, 0.071          & 86.9, 40.5, 41.4    & 8.8$\degree$ ,  8.4$\degree$      & -  & -  \\
                             &  NAOCS &  4.2$\degree $, 12.1$\degree $, 13.5$\degree $ & 0.157, 0.252, 0.168       &   -                    &   13.7$\degree$, 15.1$\degree$        & -  & -                                   \\
glasses                      & \nameshort       & \textbf{3.7$\degree $, 5.1$\degree $, 3.7$\degree $}                 &  \textbf{0.035, 0.051, 0.057}              & \textbf{87.4, 43.6, 44.5}              & \bf{4.3$\degree$ , 4.5$\degree $  }          & \textbf{2.2$\degree $ ,  2.3$\degree $}	          &      	\textbf{0.019 , 0.014}                   \\  \midrule    

\multirow{3}{*}{Oven} &  NPCS     &  1.3$\degree $, 3.5$\degree $ & 0.032,	0.049 & 75.8 , 88.5  & 4.0$\degree $      & -      &     -    \\
                      &  NAOCS     &  1.7$\degree $, 4.7$\degree $               & 0.036 ,	0.090                 &   -              & 5.1$\degree$              &       -         &   -                                 \\
                       & \nameshort    &  \textbf{1.1$\degree $, 2.2$\degree $}  &\textbf{0.030},	\textbf{ 0.046}              & \textbf{75.9 , 89.0}                          & \textbf{2.1$\degree $}               & \textbf{0.8$\degree $}      &       \textbf{ 0.024 }                    \\
                              \midrule    

\multirow{2}{*}{Washing} &  NPCS     & 1.1$\degree $, 2.0$\degree $ & 0.043 ,	0.056  & 86.9 , 88.0  & 2.3 $\degree $     & -      &     - \\
                             & NAOCS & 1.1$\degree $ ,	3.3$\degree $   &   0.072 ,	0.119              &    -             &   3.1 \degree             &   -             &               -                   \\
Machine                             & \nameshort &  \textbf{1.0$\degree $ , 1.4$\degree $}  & \textbf{0.042}, \textbf{0.053}             &\textbf{ 87.0 , 88.3}                              & \textbf{1.00 $\degree $}              & \textbf{0.7$\degree $}       &          \textbf{0.008}                   \\  \midrule    

\multirow{3}{*}{Laptop} &  NPCS     &11.6$\degree $, 4.4$\degree $          & 0.098, \textbf{0.044}        & 35.7, \textbf{93.6}     & 14.4 $\degree$  & - & - \\
                        & NAOCS     &12.4$\degree $, 4.9$\degree $         & 0.110, 0.049        & -               & 15.2 $\degree$  & -   & -\\
                        & \nameshort & \textbf{6.7$\degree $, 4.3$\degree $ }       & \textbf{0.062, 0.044}     & \textbf{41.1}, 93.0     & \textbf{9.7  $\degree$}  & \textbf{0.5$\degree $}& \textbf{0.017 }      \\ \midrule 
\multirow{3}{*}{Drawer} &  NPCS     & 1.9$\degree $, 3.5$\degree $, 2.4$\degree $, 1.8$\degree $ & 0.032, 0.038, 0.024, \textbf{0.025} & 82.8, 71.2, 71.5, \textbf{79.3}   & 0.026, 0.031, 0.046 & - & -     \\
                        &  NAOCS    & 1.5$\degree $, 2.5$\degree $, 2.5$\degree $, 2.0$\degree $ & 0.044, 0.045, 0.073, 0.054 & -                        & 0.043, 0.066, 0.048 & - & -    \\
                        & \nameshort & \textbf{1.0$\degree $, 1.1$\degree $, 1.2$\degree $, 1.5$\degree $} & \textbf{0.024, 0.021, 0.021}, 0.033 & \textbf{84.0,72.1, 71.7}, 78.6   & \textbf{0.011, 0.020, 0.030} &  \textbf{0.8$\degree $, 0.8$\degree $, 0.8$\degree $} & -    \\

\bottomrule
\end{tabular}
\caption{\label{tabel:Performance comparison} \textbf{Performance comparison  on \textit{unseen} object instances.} The categories eyeglasses, oven, washing machine, and laptop contain only revolute joints and the drawer category contains three prismatic joints.}
 \vspace{-3mm}
\label{tab:syn}
}
\end{table*}

\subsection{Experimental Setup}

\paragraph{Evaluation Metrics.}
We use the following metrics to evaluate our method. 
\begin{itemize}
\vspace{-2mm}\item[\Cdot]  \textbf{Part-based metrics.} 
For each part, we evaluate rotation error measured in degrees, translation error, and 3D intersection over union (IoU)~\cite{song2016deep} of the predicted amodal bounding box. 


\vspace{-2mm}\item[\Cdot] \textbf{Joint states.}
For each revolute joint, we evaluate joint angle error in degrees. 
For each prismatic joint, we evaluate the error of relative translation amounts. 

\vspace{-2mm}\item[\Cdot]  \textbf{Joint parameters.} For each revolute joint, we evaluate the orientation error of the rotation axis in degrees, and the position error using the minimum line-to-line distance.
For each prismatic joint, we compute the orientation error of the translation axis. 
\end{itemize}


\mypara{Datasets.}
We have evaluated our algorithm using both synthetic and real-word datasets. To generate the synthetic data, we mainly use object CAD models from \cite{wang2019shape2motion} along with drawer models from \cite{xiang2020sapien}. Following the same rendering pipeline with random camera viewpoints, we use PyBullet\cite{coumans2018} to generate on average 3000 testing images of unseen object instances for each object category that do not overlap with our training data. For the real data, we evaluated our algorithm on the dataset provided by Michel \etal \cite{michel2015pose}, which contains depth images for 4 different objects captured using the Kinect. 

\mypara{Baselines.}
There are no existing methods for category-level articulated object pose estimation. We therefore use ablated versions of our system for baseline comparison. 
\begin{itemize}
\vspace{-2mm}\item[\Cdot] \textbf{NPCS.} This algorithm predicts  part segmentation and NPCS for each part (without the joint parameters). The prediction allows the algorithm to infer part pose, amodal bounding box for each part, and joint state for revolute joint by treating each part as an independent rigid body. However, it is  not able to perform a combined optimization with the kinematic constraints.  

\vspace{-2mm}\item[\Cdot] \textbf{NAOCS.}  This algorithm  predicts part segmentation and NAOCS representation for the whole object instance. The prediction allows the algorithm to infer part pose and joint state,  but not the amodal bounding boxes for each part since the amodal bounding boxes are not defined in the NAOCS alone. Note the part pose here is defined from the NAOCS to the camera space, different from the one we defined based upon NPCS. We measure the error in the observed object scale so that it is comparable with our method.

\vspace{-2mm}\item[\Cdot] \textbf{Direct joint voting.} This algorithm directly votes for joint-associated parameters in camera space, including offset vectors and orientation for each joint from the point cloud using PointNet++  segmentation network.
\end{itemize}
Our final algorithm predicts the full \nameshort representation that includes NPCS, joint parameters, and per-point global scaling and translation value that can be used together with the NPCS prediction for computing NAOCS.

\begin{figure*}[t]
    \centering
    \includegraphics[width=\linewidth]{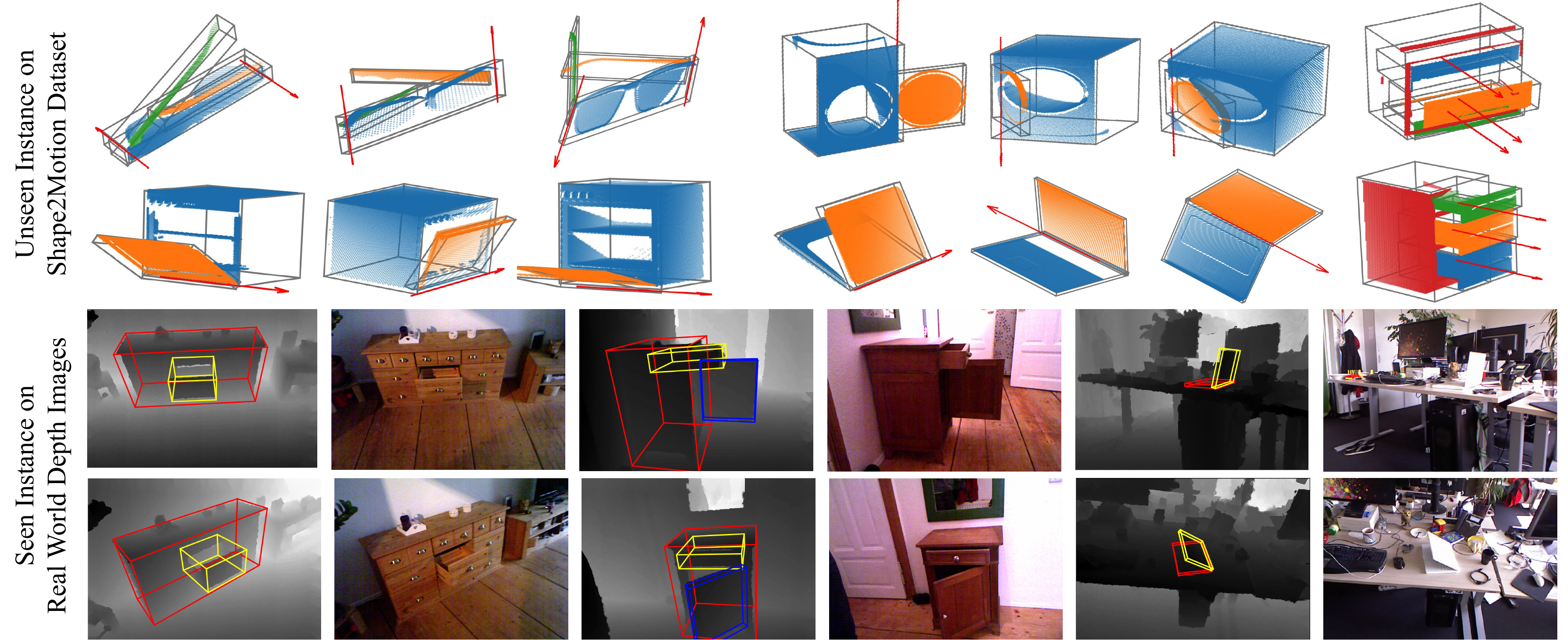}
    \caption{\textbf{Qualitative Results.} Top tow rows show test results on unseen object instances from the Shape2Motion dataset \cite{wang2019shape2motion} and SAPIEN dataset\cite{xiang2020sapien} (for only drawer category). Bottom two rows show test result on seen instances in the real-world dataset \cite{michel2015pose}. Here we visualize the predicted amodal bounding box for each parts. Color images are for visualization only. }
    \label{fig:results}
\end{figure*}

\subsection{Experimental Results}
Figure~\ref{fig:results} presents some qualitative results. Tables~\ref{tab:syn} summarizes the quantitative results. 
Following paragraphs provide our analysis and discussion of the results. 

\mypara{Effect of combined optimization.}
First, we want to examine how combined optimization would influence the accuracy of articulated object pose estimation, using both predicted joint parameters and predicted part poses.    
To see this, we compare the algorithm performance between NPCS and \nameshort, where NPCS performs a per-part pose estimation and \nameshort performs a combined optimization using the full kinematic chain to constrain the result.
The results in Table \ref{tab:syn} show that the combined optimization of joint parameters and part pose consistently improves the predict results for almost all object categories and on almost all evaluation metrics. The improvement is particularly salient for thin object parts such as the two temples of eyeglasses (the parts that extend over the ears), where the per-part based method produces large pose errors due to limited number of visible points and shape ambiguity. 
This result demonstrates that the joint parameters predicted in the NAOCS can regularize the part poses based on kinematic chain constraints during the combined pose optimization step and improve the pose estimation accuracy. 

\mypara{Joint parameters estimation.}
Predicting the location and the orientation of joints in camera space directly with all degrees of freedom is challenging. Our approach predicts the joint parameters in NAOCS since it provides a canonical representation where the joint axes usually have a strong orientation prior. We further use a voting-based scheme to reduce the prediction noise. Given joint axis predictions in NAOCS, we leverage the transformation between NAOCS and NPCS to compute corresponding joint parameters in NPCS. Based on the high-quality prediction of part poses, we will transform the joint parameters into the camera coordinate.
Comparing to a direct voting baseline using PointNet++, our approach significantly improves the joint axis prediction  for  unseen instances (Table \ref{tab:joint}).


\begin{table}[h!]
    \centering
    \setlength\tabcolsep{3px}
    \begin{tabular}{c|c|c|c}
    \toprule
  Category  & Methods & Angle error & Distance error\\
\midrule
\multirow{1}{*}{Eye-} & PointNet++   &  2.9$\degree$, 15.7$\degree$ & 0.140, 0.197\\
\multirow{1}{*}{glass}         &  \nameshort     & \textbf{2.2$\degree$}, \textbf{2.3$\degree$}  & \textbf{0.019}, \textbf{0.014}\\
\midrule
\multirow{2}{*}{Oven} & PointNet++    &  27.0$\degree$ & \textbf{0.024}\\
                      &  \nameshort     & \textbf{0.8$\degree$}  & \textbf{0.024}\\
\midrule
\multirow{1}{*}{Washing} & PointNet++   &  8.7$\degree$ & 0.010\\
        Machine          &  \nameshort     & \textbf{0.7$\degree$}  & \textbf{0.008}\\
\midrule
\multirow{2}{*}{Laptop} & PointNet++    &  29.5$\degree$ & \textbf{0.007}\\
                        &  \nameshort     & \textbf{0.5$\degree$}  & 0.017\\
\midrule
\multirow{2}{*}{Drawer} & PointNet++   &  4.9$\degree$,5.0$\degree$,5.1$\degree$ & -\\
                        &  \nameshort     & \textbf{0.8$\degree$},\textbf{0.8$\degree$},\textbf{0.8$\degree$} & -\\
    \bottomrule
    \end{tabular}
    \caption{\textbf{A comparison of joint parameters estimation.} Here PointNet++ denotes the direct joint voting baseline.}
    \label{tab:joint}
    \vspace{-5mm}
\end{table}

\mypara{Generalization to real depth images.}
We have also tested our algorithm's ability to generalize to real-world depth images on the dataset provided in \cite{michel2015pose}. The dataset contains video sequences captured with Kinect for four different object instances. Following the same training protocol, we train the algorithm with synthetically rendered depth images of the provided object instances. Then we test the pose estimation accuracy on the real world depth images. We adopt the same evaluation metric in \cite{michel2015pose}, which uses $10 \%$ of the object part diameter as the threshold to compute Averaged Distance (AD) accuracy, and test the performance on each sequence separately. Although our algorithm is not specifically designed for instance-level pose estimation and the network has never been trained using any real-world depth images, our algorithm achieves strong performance on par with or even better than state-of-the-art. On average our algorithm achieves $96.25 \%, 92.3\%, 96.9\%, 79.8\% $ AD accuracy on the whole kinematic chain of object instance laptop, cabinet, cupboard and toy train. For detailed results on each part in all the test sequences, as well as more visualizations, please refer to the supplementary material.

\begin{table*}[h!]
    \centering
    \setlength\tabcolsep{4.2px}
    \begin{tabular}{l c|c|c|c|c}
    \toprule
    Object               & \multicolumn{2}{c}{Sequence}  & Brachmann et al.\cite{brachmann2014learning} & Frank et al.\cite{michel2015pose} & ANCSH (Ours) \\
\midrule
\multirow{4}{*}{Laptop}  & \multirow{2}{*}{1}  & all     &8.9\%                 & 64.8\%                        &\textbf{94.1\%}\\
                         &                     & parts   & 29.8\% 25.1\%                 & 65.5\% 66.9\%         &\textbf{97.5\% 94.7\%}\\
    \cmidrule(lr){2-6}
                         & \multirow{2}{*}{2}  & all     & 1\%                 & 65.7\%                         &\textbf{98.4\%}\\
                         &                     & parts   & 1.1\% 63.9\%                 & 66.3\% 66.6\%         &\textbf{98.9\% 99.0\%}\\
\midrule
\multirow{4}{*}{Cabinet}  & \multirow{2}{*}{3}  & all    &0.5\%                 & \textbf{95.8\% }                       &90.0\%\\
                         &                     & parts   & 86\% 46.7\% 2.6\%                & 98.2\% 97.2\% \textbf{96.1\%}        &\textbf{98.9\% 97.8\%} 91.9\%\\
    \cmidrule(lr){2-6}
                         & \multirow{2}{*}{4}  & all     & 49.8\%                 & \textbf{98.3\%}                         &94.5\%\\
                         &                     & parts   & 76.8\% 85\% 74\%           & 98.3\% 98.7\% \textbf{98.7\%}       &\textbf{99.5\% 99.5\%} 94.9\%\\
\midrule
\multirow{4}{*}{Cupboard} & \multirow{2}{*}{5}  & all    & 90\%                 & \textbf{95.8}\%                        &93.9\%\\
                         &                     & parts   & 91.5\% 94.3\%                & 95.9\% \textbf{95.8\%}        &\textbf{99.9\%} 93.9\%\\
    \cmidrule(lr){2-6}
                         & \multirow{2}{*}{6}  & all     & 71.1\%                 & 99.2\%                         &\textbf{99.9\%}\\
                         &                     & parts   & 76.1\% 81.4\%           & 99.9\% 99.2\%        &\textbf{100\% 99.9\%}\\
\midrule
\multirow{4}{*}{Toy train} & \multirow{2}{*}{7}  & all     & 7.8\%                     & \textbf{98.1\%}                         &68.4\%\\
                         &                     & parts   & 90.1\% 17.8\% 81.1\% 52.5\%  & \textbf{99.2\% 99.9\% 99.9\%} 99.1\%        &92.0\% 68.5\% 99.3\% \textbf{99.2\%}\\
    \cmidrule(lr){2-6}
                     & \multirow{2}{*}{8}  & all    & 5.7\%                 & \textbf{94.3\%}                      &91.1\%\\
                         &                     & parts   & 74.8\% 20.3\% 78.2\% 51.2\%     & \textbf{100\% 100\%} 97\% \textbf{94.3\%}      &\textbf{100\% 100\% 100\%} 91.1\%\\
    \bottomrule
    \end{tabular}
    \caption{\textbf{Instance-level real-world depth benchmark.} While not designed for instance-level articulated object pose estimation, our algorithm is able to achieve comparable performance compare to the state-of-the-art approach and improves the performance for challenging cases such as laptops. AD accuracy is evaluated for both the whole kinematic chain(all) and different parts(parts). }
    \label{tab:real}
    \vspace{-3mm}
\end{table*}

\begin{figure}[ht]
    \centering
    \includegraphics[width=\linewidth]{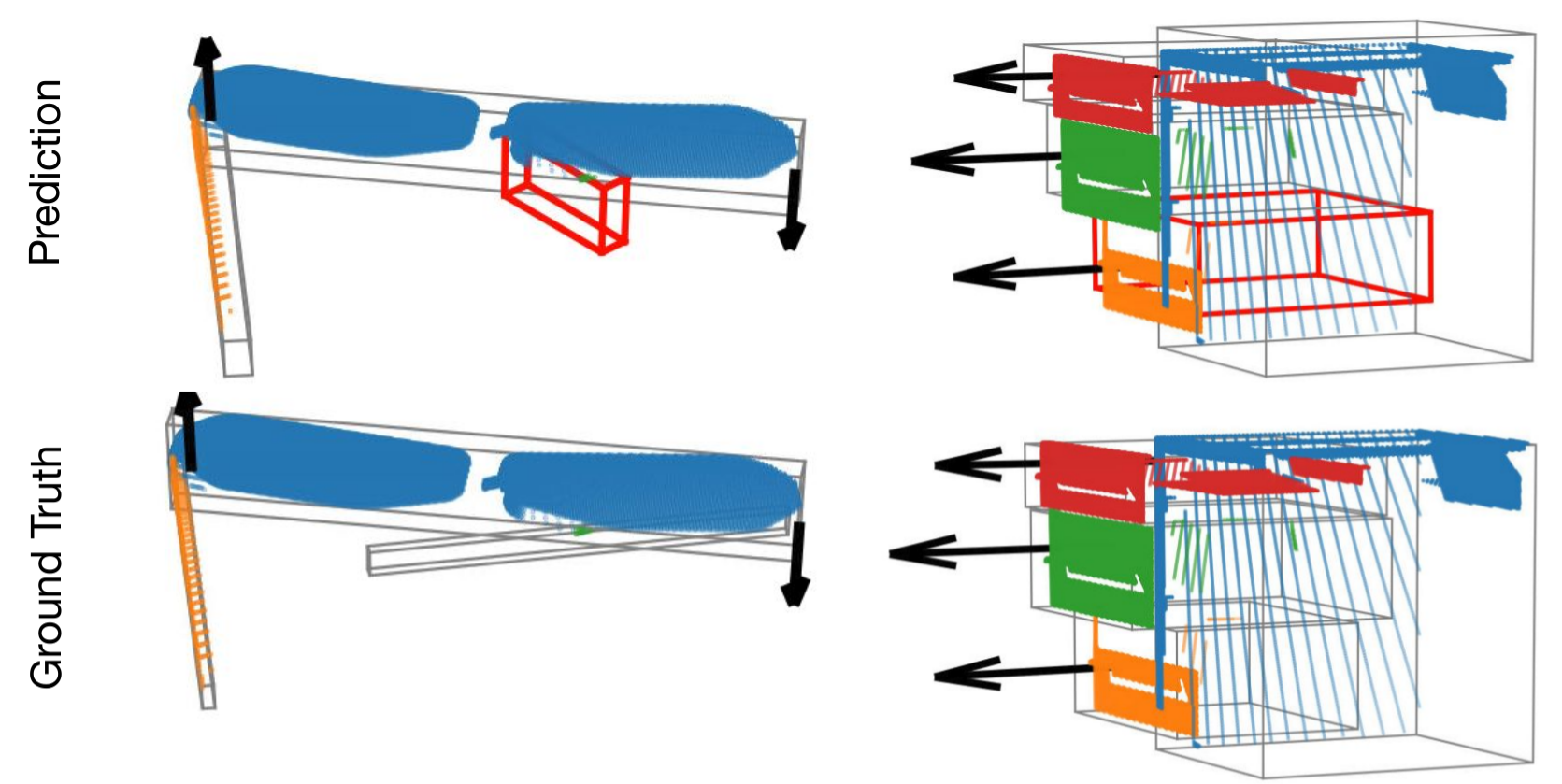}
    \caption{\textbf{Failure cases.} Left column shows failure cases on unseen eyeglasses instances, when a part is under heavy occlusion and barely visible. Right column shows the failure case on unseen drawers, when there are shape variations on parts and only the front area of the drawer is visible. The predicted drawer size is bigger than the real size. Although the box prediction is wrong, our method can reliably predict the joint state and joint parameters by leveraging the NAOCS representation.  }
    \label{fig:failures}
\end{figure}

\mypara{Limitation and failure cases.}
Figure \ref{fig:failures} shows typical failure cases of our algorithm. 
A typical failure mode of our algorithm is the inaccurate prediction under heavy occlusion where one of the object parts is almost not observed.   Figure \ref{fig:failures} shows one of such cases where one of the eye-glasses temples is almost completely occluded.
Also, under the situation of heavy occlusion for prismatic joints, there is considerate ambiguity for ANCSH prediction on the size of the heavily occluded parts, as shown in Figure \ref{fig:failures}. However, NAOCS representation does not suffer from the size ambiguity, thus leading to a more reliable estimation of the joint state (relative translation distance compare to the rest state) and joint parameters (translation axis). 

\section{Conclusion} \vspace{-2mm}
This paper has presented an approach for category-level pose estimation of articulated objects from a single depth image. 
To accommodate unseen object instances with large intra-category variations, we introduce a novel object representation, namely \namelong (\nameshort).
We further devise a deep neural network capable of predicting \nameshort from a single depth point cloud.
We then formulate articulated pose fitting from the \nameshort predictions as a combined optimization problem, taking both part pose errors and joint constraints into consideration.
Our experiments demonstrate that the \nameshort representation and the combined optimization scheme significantly improve the accuracy for both part pose prediction and joint parameters estimation.


{
\noindent \textbf{Acknowledgement:}
This research is supported by a grant from Toyota-Stanford Center for AI Research, resources provided by Advanced Research Computing in the Division of Information Technology at Virginia Tech. We thank Vision and Learning Lab at Virginia Tech for help on visualization tool. We are also grateful for the financial and hardware support from Google. 
}
\bibliographystyle{ieee_fullname}
\bibliography{main.bib}
\appendix
\section{Implementation Details}
\blfootnote{* indicates equal contributions.}
We use Tensorflow 1.10 to build our models and run the experiments for all categories. The input is uniformly sampled points from the whole back-projected depth point cloud, with points number $N$ set to 1024.  We train our model on a single Nvidia V100 GPU with batch size of 16 across the experiments. The initial learning rate is set to 0.001, with a decay factor of 0.7 every 200k steps. From the observations of our experiments, the loss will usually converge well after $>150k$ steps in less than one day. 

\section{Data generation and statistics} 
\begin{figure}[h]
    \centering
    \includegraphics[width=\linewidth]{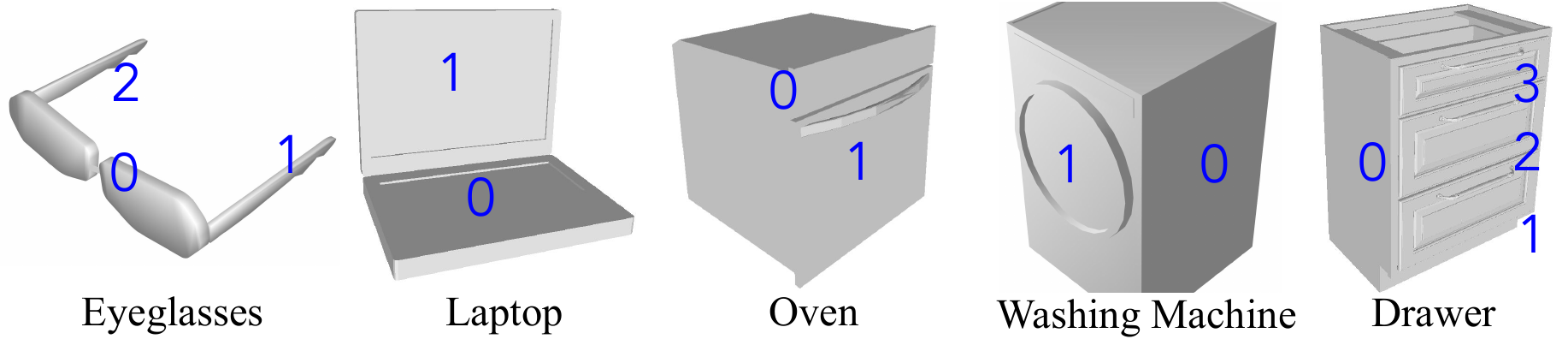}
    \centering
    \setlength\tabcolsep{3px}
    \begin{tabular}{c|c|c|c|c|c}
\toprule 
    Category  & \multicolumn{4}{c|}{Part definitions}                    & \multicolumn{1}{c}{Data statistics} \\
\midrule
   & Part 0 & Part 1 & Part 2 & Part 3 & train/test \\
\midrule
\multirow{1}{*}{Eye-} & base   & left & right & - & 39/3\\
glasses &  & temple & temple & & \\
\midrule
Oven &  base   &  door & - & - & 35/3\\
\midrule
\multirow{1}{*}{Washing} & base   &  door & - & - & 42/2 \\
Machine & & & & \\
\midrule
Laptop &  base   &  display & - & - & 78/3\\
\midrule
Drawer &  base   & lowest  & middle & top & 30/2 \\
    \bottomrule
    \end{tabular}
   \caption{\textbf{Synthetic data statistics.} We list part definitions for each object category tested in our experiments on synthetic data, together with the numbers of object instances used for training and testing.  \label{fig:part} }
\vspace{-5mm}
\end{figure}

We render synthetic depth images using the object 3D model provided in the Shape2Motion dataset \cite{wang2019shape2motion} and SAPIEN dataset \cite{xiang2020sapien}. Both datasets provide the descriptions of the object geometry and articulation information, which we leverage for generating ground truths. 
During rendering, the program automatically generates random joint states for each object instance, according to its joint motion ranges.  Then the depth images and corresponding ground truth masks are rendered from a set of random camera viewpoints. We also filter out camera poses where some parts of the object are completely  occluded. 
Figure \ref{fig:part} shows the index definitions of parts for each object category used in the main paper, together with number of object instances splitted for training and testing. 

We use real data from ICCV2015 Articulated Object Challenge \cite{michel2015pose}, which contains RGB-D data with 4 articulated objects: laptop, cabinet, cupboard and toy train. 
 This dataset provides 2 testing sequences for each object. Each sequence contains around 1000 images captured by having a RGB-D camera slowly moving around the object. Objects maintain the same articulation state within each sequence. Each part of the articulated object is annotated with its 6D pose with respect to the known CAD model. Since no training data is provided, we use the provided CAD models to render synthetic depth data, with 10 groups of random articulation status considered.
 We render object masks for the testing sequences with Pybullet\cite{coumans2018}. 

\section{Handling severe occlusion cases}
\begin{figure}[h]
 \vspace{-2mm}
    \centering
    \includegraphics[width=\linewidth]{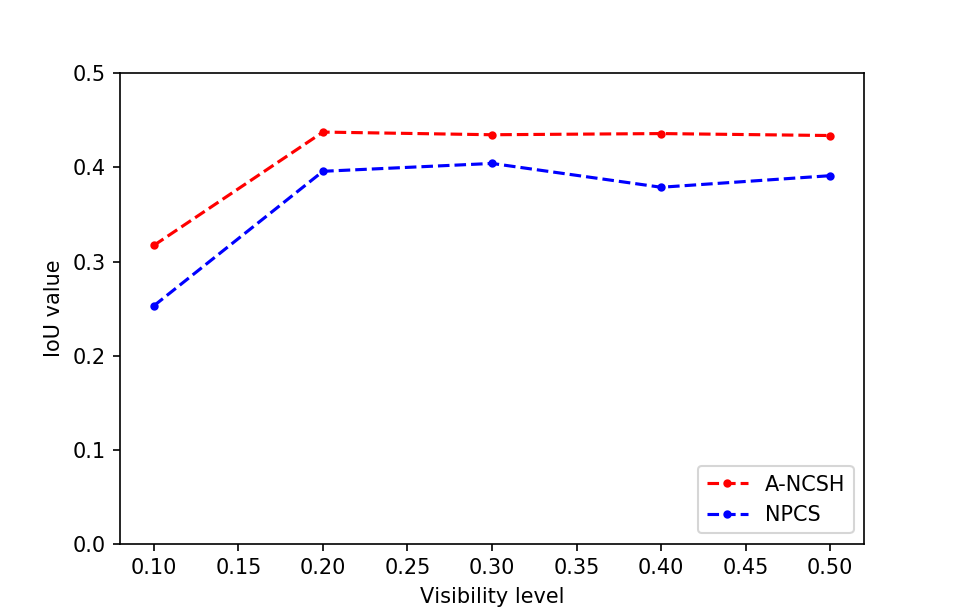}
    \caption{\textbf{ Performance under different occlusion levels.} Data is collected from part 2 of unseen eyeglasses.}
    \label{fig:occlusion}
    \vspace{-3mm}
\end{figure}

We carefully examine how ANCSH performs under different levels of occlusion. Compared to our NPCS baseline, our proposed method is still capable of improving the pose estimation under severe occlusion, as shown in Figure \ref{fig:occlusion}. The occlusion level is defined according to the ratio of visible area with respect to the total mesh surface per part.

\section{Additional results}
\begin{figure*}
    \centering
    \includegraphics[width=\linewidth]{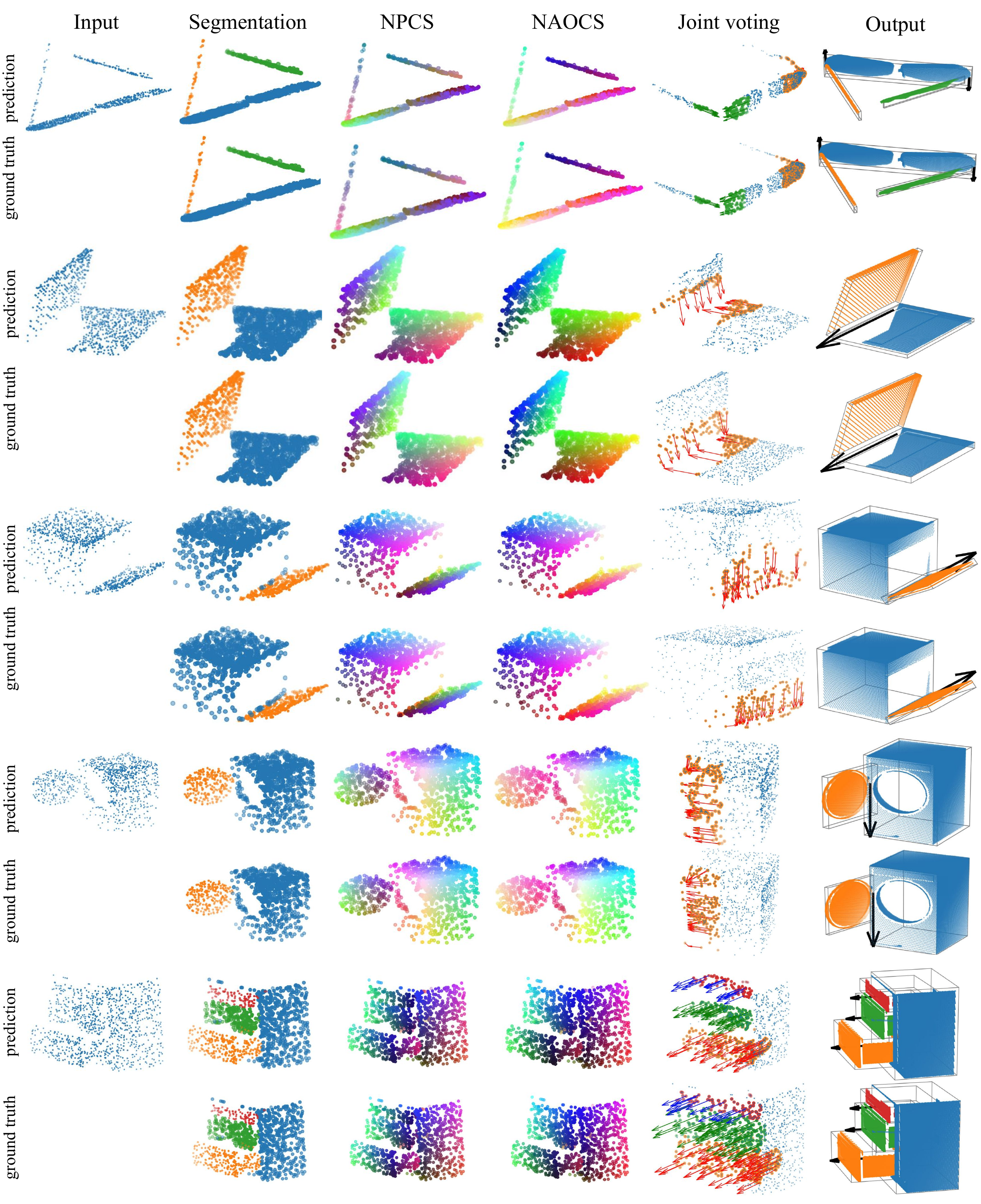}
    \caption{\textbf{Additional results on category-level synthetic dataset.} The first column shows the input point clouds; the second column shows our prediction and ground truth part segmentation mask; the third and fourth column show our prediction and ground truth NPCS and NAOCS, where the RGB channels encode the coordinates; the fifth column visualizes joint voting, where the arrows represent offset vectors to rotational hinge for revolute joints and the direction of joint axis for prismatic joints; the sixth column visualizes per part 3D bounding boxes, together with joint parameters. }
    \label{fig:syn_more}
\end{figure*}

\begin{figure*}
    \centering
    \includegraphics[width=0.95\linewidth]{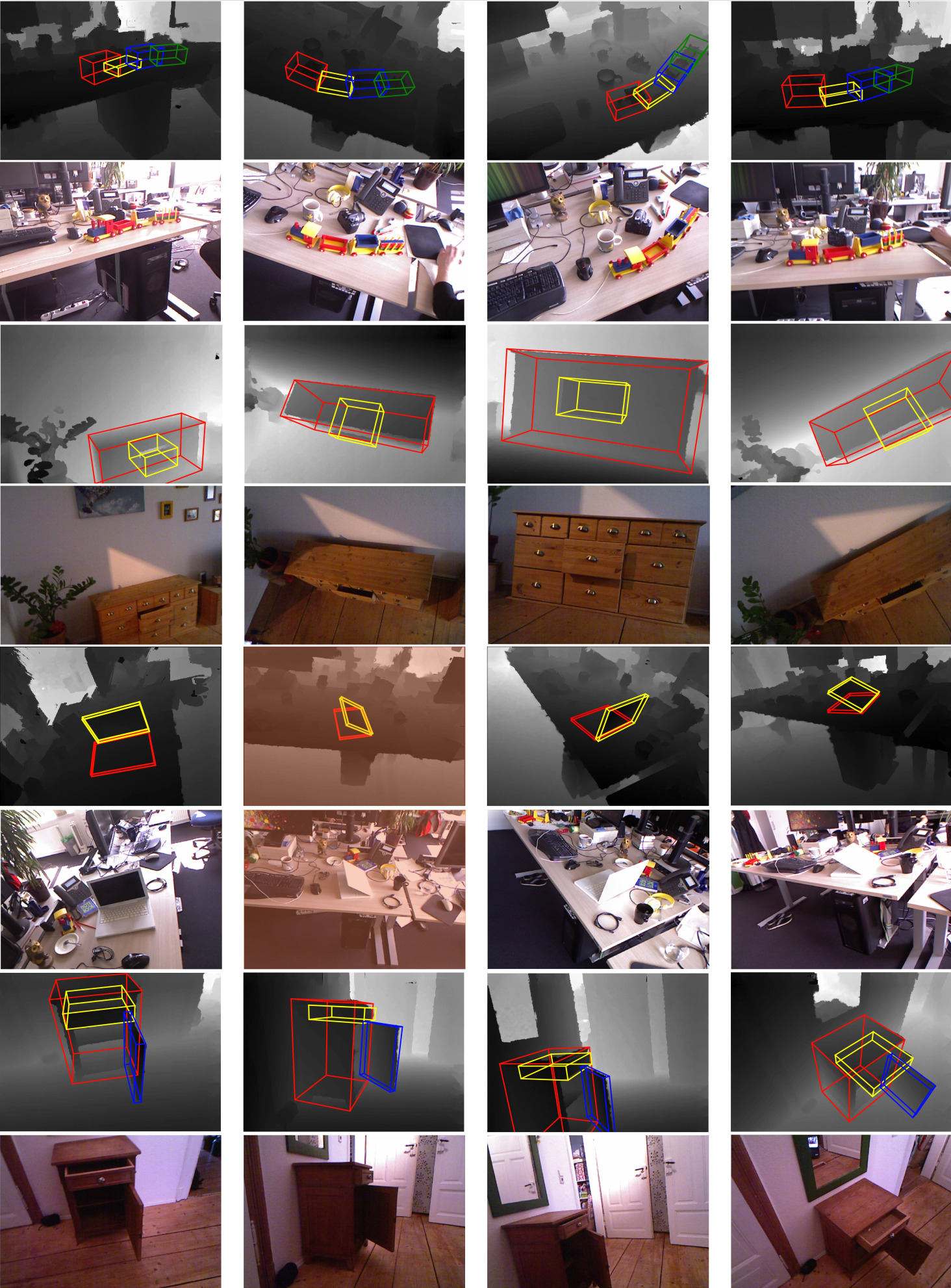}
    \caption{\textbf{Additional results on real-world instance-level depth dataset.} More qualitative results on all 4 objects from  ICCV2015 Articulated Object Challenge \cite{michel2015pose} are shown here, with toy train, cupboard, laptop, cabinet from up-pest row to lowest row in order. Only depth images are used for pose estimation, RGB images are shown here for better reference. For each object, we estimate 3D tight bounding boxes to all parts on the kinematic chain, and project the predicted bounding boxes back to the depth image. }
    \label{fig:real}
\end{figure*}

 Figure \ref{fig:syn_more} shows  additional qualitative results on the synthetic dataset. More qualitative results on real-world dataset are visualized in Figure \ref{fig:real}. 

\pagestyle{empty}

\end{document}